\title{Improving the Interpretability of fMRI Decoding\\ using Deep Neural Networks and Adversarial Robustness}

\documentclass[a4paper]{article}
\usepackage{fullpage}
\usepackage{authblk}
\usepackage{siunitx}
\usepackage[utf8]{inputenc} 
\usepackage[T1]{fontenc}    
\usepackage{hyperref}       
\usepackage{url}            
\usepackage{booktabs}       
\usepackage{amsfonts}       
\usepackage{nicefrac}       
\usepackage{microtype}      
\usepackage{algorithm}
\usepackage[noend]{algpseudocode}
\usepackage{amsmath}
\usepackage{graphicx}
\usepackage{subcaption}
\usepackage{comment}
\usepackage[numbers]{natbib}
\usepackage{lineno}
\usepackage{color,soul}

\DeclareMathOperator{\argmin}{argmin} 
\DeclareMathOperator{\argmax}{argmax}

\author[1]{{\small Patrick McClure}}
\author[2]{{\small Dustin Moraczewski}}
\author[1]{{\small Ka Chun Lam}}
\author[2]{{\small Adam Thomas}}
\author[1]{{\small Francisco Pereira}}
\affil[1]{{\small Machine Learning Team, Functional Magnetic Resonance Imaging Facility, National Institute of Mental Health, Bethesda, MD, 20892, USA}}
\affil[2]{{\small Data Science and Sharing Team, Functional Magnetic Resonance Imaging Facility, National Institute of Mental Health, Bethesda, MD, 20892, USA}}

\begin{document}
\maketitle

\begin{abstract}
While deep neural networks (DNNs) are being increasingly used to make predictions from high-dimensional, complex data, they are widely seen as uninterpretable ``black boxes'', as it can be difficult to discover what input information is used to make predictions. This ability is particularly important for applications in cognitive neuroscience and neuroinformatics. A saliency map is a common approach for producing interpretable visualizations of the relative importance of input features in prediction. However, many methods for creating these maps fail due to focusing too much on the input or being extremely sensitive to small input noise. It is also challenging to evaluate how well saliency maps correspond to the truly relevant input information, given that ground truth is not always available. 
  
In this paper, we briefly review a variety of methods for producing gradient-based saliency maps, and present a new adversarial training method we developed to make DNNs robust to input noise with the goal of improving interpretability. We introduce two quantitative evaluation procedures for saliency map methods in functional magnetic resonance imaging (fMRI), applicable whenever a model is being trained to decode some information from imaging data. We describe the rationale for the procedures using a synthetic dataset where the complex activation structure is known. We then use them to evaluate saliency maps produced for linear and DNN models performing task decoding in the Human Connectome Project (HCP) dataset. Our key finding is that saliency maps produced with different methods vary widely in interpretability, as measured by our evaluation procedures in synthetic and HCP data. Strikingly, even when linear and DNN models decode at comparable levels of performance, gradient-based saliency maps from the DNN score higher on interpretability  than maps derived from the linear model (via weights or gradient). Finally, saliency maps produced with our adversarial training method outperform those produced with alternative methods.

\end{abstract}

\section{Introduction and Related Work}

Deep neural networks (DNNs) are becoming increasingly used in cognitive and computational neuroscience for a variety of classification and regression tasks, including diagnosing diseases and predicting subject traits from functional or structural magnetic resonance imaging (sMRI, fMRI) data \citep{kietzmann2019deep,khosla2019machine}. In these areas, however, accurate predictions on new subjects are not enough. Knowing what, in input space (e.g. brain structure or brain activation), contains the information that allows the model to make its predictions is often at least as important. In neuroimaging, this information is crucial to developing a deeper understanding of the relationship between brain function and how it manifests in data. This will also be crucial for translational applications, such as modulation of brain function. The recent availability of very large imaging datasets is starting to allow the development of deep learning models that can outperform classical machine learning techniques in neuroimaging applications \citep{abrol2020hype,khaligh2014deep}. However, their adoption has been slowed, in part, by the perception that such models are inscrutable. For the specific application of {\em task decoding} -- predicting which cognitive state a subject is in -- from fMRI data, the most commonly used models are linear. To interpret these prediction models, the model weights are projected into input space (e.g. regression weights for each voxel) \citep{pereira2009machine,mensch2017learning}. However, this particular approach does not work for DNNs, given that they perform multiple layers of computation, with non-linear operations.

There have been many attempts to open up the DNN "black box" \citep{xie2020explainable}, across different areas of application. One of the most popular approaches is creating {\em gradient-based saliency maps}, visualizations of the input space highlighting the features that drive prediction. These saliency maps are functions of a prediction model {\em and} a specific input, as non-linearities in a DNN make it so that some input features can change how those, and other, input features affect the prediction. In contrast, the feature weights of linear model, which determine how input features affect the prediction, do not change for different inputs. There are a variety of methods for generating saliency maps. While they have been successfully deployed in some cases, they have two common failure modes in generating saliency maps:
\begin{enumerate}
\item maps dramatically change in the presence of even extremely small input noise 
\item maps are almost entirely a function of the specific input, and not sensitive to the learned parameters of the model 
\end{enumerate}

Both problems can be illustrated by reference to the situation where a DNN does task decoding based on activation maps. The first problem occurs when DNNs are susceptible to abrupt changes in prediction (e.g. which task) based on small input changes (e.g. tiny activation changes in a few voxels) \citep{szegedy2013intriguing,goodfellow2014explaining}. This violates the intuition that inputs that are extremely similar should have similar predictions (i.e. the model should learn a smooth function). This will then perturbs any estimate of how much a particular input feature is relevant for a prediction
\citep{kindermans2019reliability, etmann2019connection}.
Robustness, in this context, means ensuring that small changes in inputs to the DNN will not dramatically change the output. As a result, making DNNs robust to small noise has been proposed as means of increasing interpretability; however, inducing too much smoothness can lead to a decrease in accuracy in some instances \citep{tsipras2018robustness,kim2019bridging,shafahi2019adversarial}. The main approaches devised to address this rely on either smoothing the input gradients of non-robust DNNs (e.g. SmoothGrad \citep{smilkov2017smoothgrad}) or making DNNs robust to input noise, namely using random and adversarial input noise during training. We include all three of these methods in the experiments in this paper. To the best of our knowledge, adversarial training for improving interpretability has not been used with neuroimaging data.The second problem occurs when methods produce saliency maps that mostly reflect the specific input (e.g. they are a function of activation being present or absent), rather than what the model learned (e.g. which presence of activation distinguishes some classes from others). This can be demonstrated by showing that saliency maps are similar for both a trained DNN and one where parameters were randomly initialized, which would indicate that the maps do not reflect the information the model uses to make a prediction. \cite{adebayo2018sanity} shows this is the case for methods such as
multiplying the gradient by the input, guided back-propagation, guided GradCAM, and integrated gradients all over-focus on the specific input. Note that other methods, such as LRP and DeepLift, which is used in SHAP \citep{lundberg2017unified}, are equivalent to the gradient times input method, for ReLU networks with zero baseline and no biases \citep{ancona2018towards}. Due to this, and the popularity of LRP and DeepLift, we will, in all experiments, use the gradient times input method as a representative technique for the methods that over-focus on the input when generating saliency maps. The methods above have been used to study DNNs trained on neuroimaging data, with \cite{wang2018decoding} using guided back-propagation, \cite{thomas2019analyzing} using LRP, and \cite{ismail2019input} using standard gradients. Note that \cite{adebayo2018sanity} found that the input gradient and SmoothGrad saliency maps did change in response to randomizing DNN weights.
%


A different set of issues in using gradient-based saliency maps stems from the difficulty of getting a quantitative measure of the quality of a map. Intuitively, a saliency map is good if it highlights the features that help the DNN distinguish between classes; this is function of both the data and the specific prediction the DNN makes. In the context of a DNN for fMRI decoding, the saliency map for a particular task should show the locations where the presence of activation is associated with that {\em specific} task alone and the locations where activation is {\em shared} between that and some (but not all) of the other tasks. Additionally, the saliency map for a task should indicate the locations where activation indicates some other task, since if activation is present in those locations it is evidence of another task.

In the neuroimaging literature, the methods used to evaluate gradient-based saliency maps have focused on comparing to reference images, often ignoring negative gradients (e.g. \cite{wang2018decoding} compared to general linear model (GLM) voxelwise activation image for the tasks and \cite{thomas2019analyzing} compared to NeuroSynth relevance maps \citep{yarkoni2011neurosynth} for the different tasks). Note that reference images such as those are population wide explanations, and comparing to them ignores the fact that DNN gradients are input-specific (i.e. a function of each {\em individual} input image and the prediction task), instead of population-wide (i.e. a function of the prediction task alone) as in linear models.

In the machine learning literature, \cite{tsipras2018robustness} compares the saliency maps of non-robust and robust DNNs, but the gradient quality is only evaluated using subjective human visual inspection. \cite{kim2019bridging} use both visual inspection and Remove and Retrain (ROAR) and Keep and Retrain (KAR) \cite{hooker2019benchmark}. These evaluation methods require removing or keeping, respectively, the most salient features and retraining. This, however, is extremely computationally expensive and can destroy information about interaction between features. Also, \cite{kim2019bridging} found that the previously mentioned methods that over-emphasise the input, per \cite{adebayo2018sanity}, do well on this evaluation.  \cite{etmann2019connection} considers the link between adversarial training and interpretability, but uses alignment with the input as a measure of interpretability, which the findings of \cite{adebayo2018sanity} call into question. 

In this paper, we propose two quantitative evaluation criteria for interpretability in fMRI decoding, and use them to compare the performance of different saliency map methods. These criteria take advantage of the fact that, for certain datasets, ground truth saliency maps can be computed. To illustrate this rationale, and compare the quality of the various methods, we demonstrate these evaluations in a synthetic dataset that mimics brain activation with known information structure.  We will show that, under those quantitative criteria, saliency maps produced with different methods vary widely in interpretability. We also apply several saliency-based interpretability methods to linear and DNN models trained to perform fMRI task decoding using the Human Connectome Project (HCP) dataset \citep{van2013wu}. HCP data allow us to estimate ground truth saliency maps for each subject, thereby enabling us to test the degree to which the method can combine information in the subject-specific input and information extracted by the network across all the subjects in the training set. Our experiments evaluate a selection of methods from those described above, as well as a new adversarial training method we introduce in this paper. Our experimental results indicate that making DNNs robust to small input noise, via adversarial training, can greatly improve the interpretability of saliency maps, without adversely affecting decoding performance. Finally, we provide code for reproducing our results in synthetic and real datasets.\footnote{\textnormal{https://github.com/patrick-mcclure/adversarial\_training\_for\_fmri\_decoding}}

\section{Methods}

\subsection{Prediction Models}

\paragraph{Multinomial Regression}

In multinomial regression, each class (e.g decoded task), $y_i$, has a weight for each input feature (e.g. voxel), resulting in a weight matrix $W$ where row $i$ contains the weight vector ${\bf w_i}$ for class $i$. These weights are then multiplied by the input features, ${\bf x}$, to yield a vector ${\bf z}= W {\bf x}$, with one logit $z_i$ per class $i$. A softmax function, $\sigma(z_i) = \frac{\exp(z_i)}{\sum_j \exp(z_j)}$, is then applied to the resulting vector, computing $\sigma(W{\bf x})$. The output of the softmax is a probability distribution over classes that is parameterized by $W$, $p_W(y|{\bf x})$. When incorporating L2 regularization, training becomes maximum a posteriori (MAP) estimation, $\argmax_W p_W(y|{\bf x}) p(W)$, with a Gaussian weight prior, $p(W)$.

\paragraph{Neural Networks}

In many cases, a multinomial regression model operating on input features cannot separate all of the classes. Deep neural networks (DNNs) are models that can learn to compute new features that are non-linear functions of the original input features, yielding a new feature space where classes can be separated. A DNN consists of functions ("layers"). A layer, $l$, takes the form of multiplying the input to the layer, ${\bf x}_l$, by a weight matrix, $W_l$, and then applying a non-linearity, $f_l$, resulting in new features $f_l(W_l {\bf x}_l)$. Layers are applied sequentially, with the first layer operating on the input image, the second layer operating on the output of the first, and so on until the layer before the last. For multi-class classification, the final layer usually computes a softmax non-linearity, similar to multinomial regression, producing $p_W(y|{\bf x})$. 


\subsection{Quantitative Evaluation of Interpretability}

\paragraph{Interpreting using Input Gradients}
\label{sec:interpretability.gradients}

One definition of the interpretability of a prediction model is the degree to which it is possible to understand the relationship between the input and output variables learned by that model. For linear regression, this is usually done by looking at the learned weights, as they directly reflect how much changing an input variable would change the output variable (i.e. the gradient of the output with respect to the input). However, linear regression weights are not always straightforward to interpret, as discussed in \cite{kriegeskorte2019interpreting}. In our classification setting, interpretability translates into being able to understand what, in the input space, allows the model to distinguish classes (i.e. how changing the input variables changes the probability of predicting each class). Formally, this is the gradient with respect to the {\em input variables}, $\nabla_{\bf x} p_W(y|{\bf x})$. This is different from the gradient with respect to the weights, the {\em learned parameters} of the model, $\nabla_W p_W(y|{\bf x})$. This gradient is computed during the training of a DNN, in gradient-based optimization methods,  to find parameters that make the training data more likely given the model. Note that the gradients for DNNs are dependent on the inputs; this is in contrast to linear regression, which has the same gradient for {\em every} input. Note that different models can produce different probabilistic predictions, and therefore different gradient maps, and still have the same accuracy. This is because the accuracy depends only on what class is predicted to be most likely, but does not take into account exactly how likely that, or other, classes are according to a model's prediction.

\begin{figure}[htb]
  \centering
    \includegraphics[width=\textwidth]{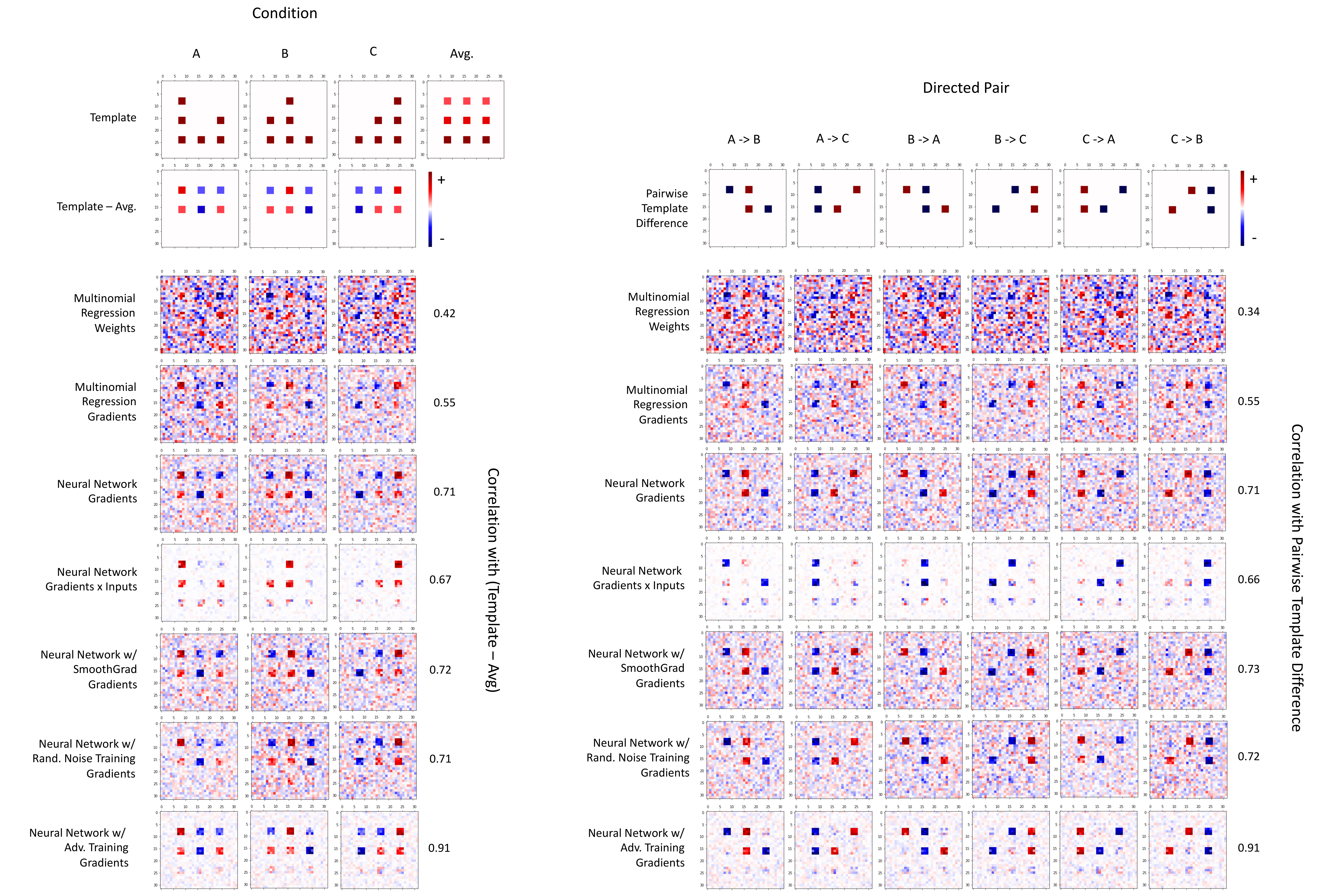}
    \begin{subfigure}[b]{0.46\textwidth}
    \captionsetup{justification=centering}
    \caption{}
  \end{subfigure}
  \begin{subfigure}[b]{0.53\textwidth}
    \captionsetup{justification=centering}
    \caption{}
  \end{subfigure}
  \caption{{\bf Visualization of saliency maps for synthetic test data predictions, for the two quantitative evaluations we introduce.} {\bf a:} The average saliency maps for the correct class and their Pearson correlations with the corresponding "template - average" map. {\bf b:} The average saliency maps for other classes and their correlations with the corresponding "pairwise template difference" map.}
  \label{fig:sim}
\end{figure}

\paragraph{Quantitative Criteria for Evaluation of Interpretability} 
\label{sec:synth-methods}

In this section, we propose two quantitative evaluation criteria for the interpretability of gradient-based saliency maps for a classification model. We will use them to compare a variety of methods for producing saliency maps, which will be introduced in the rest of this section. The experimental results are presented in Section~\ref{sec:experiments}, using both synthetic and real data. Given that it will be easier to describe the evaluation criteria with reference to illustrative examples, we will introduce the synthetic dataset at this point.

The synthetic dataset is comprised of 2D images that are simplified analogues of brain activation maps. The "brain" in these maps can be in one of three classes: A, B, or C. Each class has a characteristic template of activation, made up of 6 activation regions, shown in row 1 of Figure~\ref{fig:sim}a, together with the average of the three templates. Each class template (e.g. A) has an active region that is completely {\em specific} to it, and appears in no other class. In addition, it has two other active regions that are {\em shared} with one other class (e.g. AB and AC). Finally, it has three active regions that are present in all classes (e.g. ABC). 

Our first evaluation criterion aims to answer the question, "What information in the input space allows the model to make a correct prediction?" For activation maps, it is whatever activation departs from the average activation across all classes. Intuitively, this means that any asymmetry in the presence of activation for different classes provides information that the prediction model can exploit. We quantify this by computing the pixelwise correlation between a saliency map for the correct class of an image and the "template - average" image (i.e. the difference between the correct class template and the average of the three templates) for each test image. We will call this "template - average" image a {\em class informative activation map} (row 2 of Figure~\ref{fig:sim}a). The values in this map should also reflect the relative importance of class-specific versus shared activation, in terms of carrying information about the class. This map will be the ground truth for the gradient-based saliency maps produced for each class by the various methods we consider. These are shown in rows 3-9 of Figure~\ref{fig:sim}a, together with their average Pearson correlation with the ground truth across test examples, and will be discussed in Section~\ref{sec:experiments}.

Our second evaluation criterion aims to answer the question, "What information in the input space corresponds to the important differences between any two classes?" The answer to this, given an image of a particular class (e.g. A), are the changes in activation that would be needed to transform it into an image of another class (e.g. B). We will call this the {\em pairwise class activation difference map}, shown in row 1 of Figure~\ref{fig:sim}b. The saliency map for predicting class B, with a input from class A, should approximate this map, and can be calculated suing the fact that a gradient-based saliency map depends on both the input image and the prediction target of interest. We quantify this by computing the pixelwise correlation between a saliency map for a class (e.g. B), given an image of another class (e.g. A), and the pairwise class activation difference map (e.g. B-A). This map will be the ground truth for the gradient-based saliency maps produced for each class transformation, by the various methods we consider. These are shown in rows 2-8 of Figure~\ref{fig:sim}b, together with their average correlation with the ground truth across test examples, and will be discussed in Section~\ref{sec:experiments}.

\subsection{Methods to Compute Gradient-based Saliency Maps}

\paragraph{Multinomial Regression Weights and Gradients}

For multinomial regression, two options for producing a saliency map are using the learned weights for each input feature, or computing the gradient of the output with respect to the input. We include both of these in all experiments, and show the resulting gradients and weights for the synthetic data in rows 3/4 and 2/3 of Figure~\ref{fig:sim}a and b, respectively.

\paragraph{DNN Gradients and Gradients Times Input}

When using a DNN, the two most common options for visualizing the input features that drive prediction are gradient-based saliency maps, and gradient-based saliency maps multiplied by their input. As discussed earlier, the latter is equivalent to several other commonly used methods. We include both of these methods in all our experiments, and show the resulting saliency maps for the synthetic data in rows 5/6 and 4/5 of Figure~\ref{fig:sim}a and b, respectively.

\paragraph{SmoothGrad}

One popular method for dealing with the lack of robustness in DNN gradients is SmoothGrad \cite{smilkov2017smoothgrad}, which estimates the gradient $\nabla_{\bf x} p_W(y|{\bf x})$ around an input ${\bf x}$ as $\mathop{\mathbb{E}}_\delta[ \nabla_x p_W(y|{\bf x}+\delta)]$, where $p(\delta)=\mathcal{N}(\delta;\mathbf{0},\,I\sigma_\delta^{2})$. In practice, Monte-Carlo sampling is used to estimate this expectation by adding random noise $\delta$ -- sampled from $p(\delta)$ -- to an input, calculating the input gradient for each of these noisy inputs, and then averaging over the gradients produced for these different noisy samples. Several papers have reported that this improves the interpretability of DNN gradients \citep{adebayo2018sanity,smilkov2017smoothgrad}, although there are no formal guarantees on this leading to robust gradients. We include SmoothGrad in all experiments, and show the resulting saliency maps for the synthetic data in rows 7 and 6 of Figure~\ref{fig:sim}a and b, respectively.

\paragraph{Random Noise and Adversarial Training}
A common way of increasing model robustness to input noise is to add noise to inputs during the training process, so that the DNN encounters different variations of a given training example, all of which must lead to the same prediction. The input noise during training is often random noise such that the produced training examples variations are centered at the original training examples, such a zero-mean Gaussian noise or zero-centered spherical noise.  We include models trained with random noise in all experiments, and show the resulting saliency maps for the synthetic data in rows 8 and 7 of Figure~\ref{fig:sim}a and b, respectively.

It is possible to train with noise that more effectively increases a DNNs robustness by using {\em adversarial training}, where the noise is determined by calculating the input change that would alter the output the most, rather than randomly.
In the original adversarial training papers, the adversarial noise was computed using a gradient step for the input gradient at a real input example, using the fast gradient sign method \citep{szegedy2013intriguing,goodfellow2014explaining}. Recently, projected gradient descent (PGD) has been used to improve adversarial training by performing multiple gradient steps when generating adversarial noise \citep{madry2017towards}. For PGD, the goal is to find the adversarial noise that will either optimally decrease the probability of the correct class of a training input (a.k.a. non-targeted adversarial noise) or optimally increase the probability of an incorrect class (a.k.a. targeted adversarial noise) while staying close to the input example. A common definition of staying close to the input example is restricting the adversarial noise, $\delta$, to have a small L2-norm (i.e. keeping the Euclidean distance between the input example and the adversarial example small). Generating a non-targeted, L2 adversarial example using PGD is achieved by optimizing $\argmin_{\delta_{({\bf x},y)}} \: p_W(y|{\bf x}+\delta_{({\bf x},y)}) \; \textrm{s.t.}\: \| \delta_{({\bf x},y)} \|_2 \leq \epsilon$. 

Adversarial noise can be found by performing this optimization, and then adding it to the respective input in order to generate an adversarial example. These adversarial examples can then be used to calculate the parameter gradients for one parameter update step. For relatively small models, such as our synthetic data experiment, this can work very well. However, this is computationally expensive and can be prohibitive if using large DNNs. Recently, \cite{shafahi2019adversarial} proposed simultaneous generation of adversarial examples, by optimizing the adversarial noise, and updating of the model parameters, thus significantly speeding up adversarial training. We developed a version of this method that uses example-specific L2 adversarial noise, instead of universal L$_\infty$ noise (see Algorithm \ref{adv}). L2 noise was chosen because it has been suggested to have improved interpretability in comparison to $L_{\infty}$ noise \citep{tsipras2018robustness}, and has less of an accuracy-interpretability tradeoff than $L_{\infty}$ noise \citep{kim2019bridging}.  We include models trained with our adversarial training method e in all experiments, and show the resulting saliency maps for the synthetic data in rows 9 and 8 of Figure~\ref{fig:sim}a and b, respectively.

\begin{algorithm}
\caption{m-step Minibatch Adversarial Training}
\label{adv}
\begin{algorithmic}[1]
\Require {Set of training minibatches $\mathcal{D}$, Model parameters $W$, noise bound $\epsilon$, learning rate $\tau$, hop steps $m$}
\While {not converged}
    \For {minibatch $B \in \mathcal{D}$}
        \For {$({\bf x},y) \in B$}
            \State {$\delta_{({\bf x},y)} \leftarrow {\bf 0}$ \Comment{Initialize adversarial noise}}
        \EndFor
        \For {$i = 1,...,m$}
            \State {$g_{W} \leftarrow \mathbb{E}_{({\bf x}, y) \in B}\left[\nabla_{W} \log p_W(y|{\bf x}+\delta_{({\bf x},y)})\right]$ \Comment{Get parameter gradients}}
            
            \State {$W \leftarrow W + \tau g_{W}$ \Comment{Update parameters}}
            
            \For {$({\bf x},y) \in B$}
                \State {$\left.g_{adv} \leftarrow \nabla_{{\bf x}} p_W(y|{\bf x}+\delta_{({\bf x},y)})\right]$\Comment{Get input gradients}}
                \If {$\| g_{adv} \|_2 > 0$}
                    \State {$\nu \sim  U(0,\epsilon)$\Comment{Sample adversarial step size}}
                    \State {$\delta_{({\bf x},y)} \leftarrow \delta_{({\bf x},y)}- \nu g_{adv} / \| g_{adv} \|_2$\Comment{Update adversarial noise}}
                \EndIf
                \If {$\| \delta_{({\bf x},y)} \|_2 > \epsilon$}
                    \State {$\delta_{({\bf x},y)} \leftarrow \epsilon \delta_{({\bf x},y)} /\| \delta_{({\bf x},y)} \|_2  $\Comment{Rescale the adversarial noise}}
                \EndIf
            \EndFor
        \EndFor
    \EndFor
\EndWhile
\end{algorithmic}
\end{algorithm}

\paragraph{Practical Considerations when using Input Gradients}

Unlike the derivative of a logit, the derivative of the probability, $\nabla_{\bf x} p_W(y|{\bf x})$, is dependent on the confidence of the network. This can be an issue if a model is overconfident, since predicted probabilities very close to 0 or 1 will make the derivatives be very close to 0. To prevent this, we perform temperature scaling for the evaluated models, as suggested by \cite{guo2017calibration}. In temperature scaling, a temperature parameter, $t$, is introduced into the softmax function, $\sigma(z)_i = \frac{\exp(z_i/t)}{\sum_j \exp(z_j/t)}$. The temperature parameter can then be set by taking an already trained model, fixing its weights, and finding the $t$ that results in the lowest negative log-likelihood on the validation data. This will result in a model whose predicted probabilities better match the data, since the negative log-likelihood penalizes being confident on incorrect predictions. 

An additional complication of using gradients of non-linear models for interpretation stems from the fact the gradients are dependent on the inputs. An advantage of an input-specific gradient is that one can compute which input changes are important for a given example (e.g. what would transform the brain of a particular patient with a disorder into a healthy one). Conversely, a potential disadvantage is that these input-specific gradients may not directly correspond to important population-level changes.

\section{Experiments}
\label{sec:experiments}
\subsection{Synthetic Data}


\paragraph{Data} As described in Section~\ref{sec:synth-methods}, the synthetic dataset contains 2D images that are simplified analogues of brain activation maps in one of three classes: A, B, or C. Each class activation template (top row of Figure~\ref{fig:sim}a) contains $32 \times 32 = 1024$ pixels. Each activation region is a $3 \times 3$ pixel square, so $\sim 7\%$ of the "brain" contains activation. Pixels belonging to an activation region have a value of 1, background pixels have a value of 0. We generated 1000 examples of each class for train, validation, and test sets, respectively. Each example was produced by taking the template for the corresponding class and corrupting it with pixelwise independent Gaussian noise ($\sigma=0.5$).

\paragraph{Prediction Models and Results} We trained both an L2-penalized multinomial linear regression and a NN with L2 regularization, with a 20 unit fully connected layer with ReLU non-linearities and a 7 unit fully connected layer with a softmax non-linearity, to predict which of the three classes (i.e. A, B, or C) generated an image. Models were trained with full-batch Adam and the NN was trained with and without 3-step PGD adversarial training with $\epsilon=1$. To test if adversarial noise was particularly helpful, we also trained an NN with additive uniform random noise within the same $\epsilon$-sphere as used in adversarial traning. For SmoothGrad, $\sigma_\delta = 0.3$, and was set to maximize interpretability on the validation data. Each model was able to perfectly classify the test data.  

\paragraph{Class Informative Activation}

The first criterion introduced in Section~\ref{sec:synth-methods} evaluates how well a gradient-based saliency map for a class, generated from an input image with that class, identifies class-informative activation. The ground truth for this is the class informative activation map (the difference between the class template and the average template across all classes), shown in row 2 of Figure~\ref{fig:sim}a. 
Each gradient-based saliency map method we consider produces, for each example, a map that depends on its class. The average of these across test examples is shown in rows 3-9 of Figure~\ref{fig:sim}a. The evaluation measure for each test example map is its pixelwise correlation with the class informative activation map, yielding a sample of correlation values. We use a two-sided paired t-test to compare whether the average correlation for any two methods is different, with Bonferroni correction to account for all of the pairwise comparisons being carried out.

We found that the multinomial regression {\em gradients} significantly increased the correlation with the class informative activation maps, compared to the multinomial regression {\em weights} ($p<0.01$, corrected). This interpretability improvement was expected, because the weights for one class do not take into consideration the softmax and the weights for the other classes. We also found that the NN gradients significantly outperformed the multinomial regression gradients ($p<0.01$, corrected) and that using either additive random noise during training or SmoothGrad did not significantly increase the correlation of the NN gradients with the class informative activation maps ($p>0.05$, corrected). The gradient times input maps had increased correlation with the class informative activation maps ($p<0.01$, corrected), but had decreased correlation compared to all of the other methods ($p<0.01$, corrected). While the gradient times input maps have very little background noise, the method eliminates negative gradients in regions not activated by the corresponding class template and exaggerates gradients in regions active in the corresponding template that do not have large gradients. Adversarial training significantly increased the correlation between the gradient and class informative activation maps compared to the NN gradients ($p<0.01$, corrected), the gradients for the NN trained with additive random noise ($p<0.01$, corrected), and  the SmoothGrad gradients ($p<0.01$, corrected).

\paragraph{Pairwise Class Activation Differences}

The second criterion introduced in Section~\ref{sec:synth-methods} evaluates how well the saliency map for a class B, given an input from class A, identifies the changes in activation that would be needed to transform the input from class A to class B. The ground truth for this, in the synthetic data, is the pairwise class activation difference map, the difference between the template images for classes B and A. This is shown, for every possible transition, in the first row of Figure~\ref{fig:sim}b.

We applied this approach to the saliency maps produced by all of the methods, as shown in rows 2-8 in Figure \ref{fig:sim}b. For each test example, each method produces a saliency map indicating what it would require to transform the example to a given class. Each of those saliency maps is then correlated with the corresponding pairwise class activation difference map, yielding an average correlation value across possible classes; across test examples, this results in a sample of correlation values. For any two methods, we used a two-sided paired t-test to determine if their average correlation was different. We used Bonferroni correction to account for all of the pairwise comparisons. We found that the multinomial regression gradients increased the correlation with the task activation difference maps, compared to the multinomial regression weights ($p<0.001$, corrected). We also found that the NN gradients significantly outperformed the multinomial regression gradients ($p<0.01$, corrected). We also found that training with additive random noise decreased the correlation ($p<0.01$, corrected) and that using SmoothGrad did not change the correlation of the NN gradients with the task activation difference maps ($p<0.05$), corrected). While the gradient times input maps have very little background noise, the method eliminates positive gradients in regions not activated by the corresponding class template and exaggerates gradients in regions active in the corresponding template that do not have large gradients. Adversarial training lead to a large increase in correlation between the gradient and task activation difference maps compared to both the NN gradients ($p<0.01$, corrected), the NN trained with additive random noise ($p<0.01$, corrected), and the SmoothGrad gradients ($p<0.01$, corrected).

\subsection{fMRI data}

\begin{table}
\centering
\caption{The contrast to task mapping used to create labels for the HCP data. Note that, in the absence of a relevant control condition, we used the average of the condition versus baseline contrasts.}
\begin{tabular}{l|l}
{\bf HCP Task} & {\bf Contrasts or Conditions Used} \\ \hline
working memory &  2-BACK versus 0-BACK contrast \\
gambling & average of REWARD and PUNISH versus baseline\\
motor & average of LF,LH,RF,RH,T versus baseline\\
language & MATH versus STORY contrast \\
social & TOM versus RANDOM contrast \\
relational & REL versus MATCH contrast \\
emotion & FACES versus SHAPES contrast
\end{tabular}
\label{tab:tasks}
\end{table}

\paragraph{Data}

For our main evaluation, we used volumetric task fMRI activation maps from the Human Connectome Project 1200 dataset (HCP1200) distribution, for the 965 subjects for which they were produced. We randomly split subjects into groups of 788 (training), 92 (validation), and 95 (test), taking care to keep siblings within the same set, and assigning the largest families to the training set. For each subject, HCP1200 contains activation maps for several conditions within each of seven tasks (described in \cite{barch2013function}): working memory, gambling, motor, language, social, relational, and emotion. In our experiments, each task is one class, with the example being the activation map contrasting the main task condition with the control condition (see \cite{barch2013function} and Table~\ref{tab:tasks}, exceptions were made for gambling, given that activation in the two conditions was extremely similar, and motor, as there was no clear control). The input ${\bf x}$ and output $y$ pairs used in training/testing were created by individually z-scoring these activation maps, and labelling them with their respective task.

\begin{table}[h]
\centering
\caption{The 3D convolutional neural network architecture used for HCP fMRI task decoding.}
\label{Arch}
\begin{tabular}{c|c|c|c|c|c}
{\bf Layer} & {\bf Filter} & {\bf Stride} & {\bf Padding} & {\bf Non-linearity} & {\bf Pooling} \\
\hline
1 & $32\textnormal{x}3\textnormal{x}3\textnormal{x}3$ & 1 & 1 & ReLU & $2\textnormal{x}2\textnormal{x}2$ Avg. \\
2 & $32\textnormal{x}3\textnormal{x}3\textnormal{x}3$ & 1  & 1 & ReLU & $2\textnormal{x}2\textnormal{x}2$ Avg. \\
3 & $64\textnormal{x}3\textnormal{x}3\textnormal{x}3$ & 1  & 1 & ReLU & $2\textnormal{x}2\textnormal{x}2$ Avg. \\
4 & $64\textnormal{x}3\textnormal{x}3\textnormal{x}3$ & 1  & 1 & ReLU & $2\textnormal{x}2\textnormal{x}2$ Avg. \\
5 & $64\textnormal{x}3\textnormal{x}3\textnormal{x}3$ & 1  & 1 & ReLU & $2\textnormal{x}2\textnormal{x}2$ Avg.   \\
6 & $7\textnormal{x}57768256$ & -  & - & Softmax & - \\
\end{tabular}
\end{table}

\paragraph{Prediction Models and Results}

For classifying task from HCP 3D activation maps, we trained an L2-penalized multinomial linear regression model, a 3D CNN with L2 regularization, and a 3D CNN with 4-step minibatch adversarial training (Algorithm \ref{adv}). The CNN uses $3\times3\times3$ filters, with 5 convolutional layers (32, 32, 64, 64, 64 filters and ReLU nonlinearities) and one final softmax layer (Table \ref{Arch}). These were all trained using Adam \citep{kingma2014adam} and a batch size of 32. The CNNs were also trained with weight normalization \citep{salimans2016weight}. Hyperparameters, such as the L2 coefficient and adversarial noise bound, were set using the validation data. For SmoothGrad, $\sigma_\delta = 2.0$ was set to maximize interpretability on the validation data. For adversarial training, the adversarial noise bound, $\epsilon$, was set to 95, which corresponds to a noise of 0.1 (i.e. 10\% of a standard deviation) per voxel. The test set accuracies were 97.43\%, 98.39\%, 98.88\%, and 98.07\% for the linear model, the CNN, the CNN trained with additive random noise, and the CNN with adversarial training, respectively. These accuracies were not significantly different ($p>0.05$, corrected), per paired t-tests across test subjects.

\begin{figure} [h]
  \centering
  \begin{subfigure}[b]{0.4\textwidth}
    \centering
    \includegraphics[width=\textwidth]{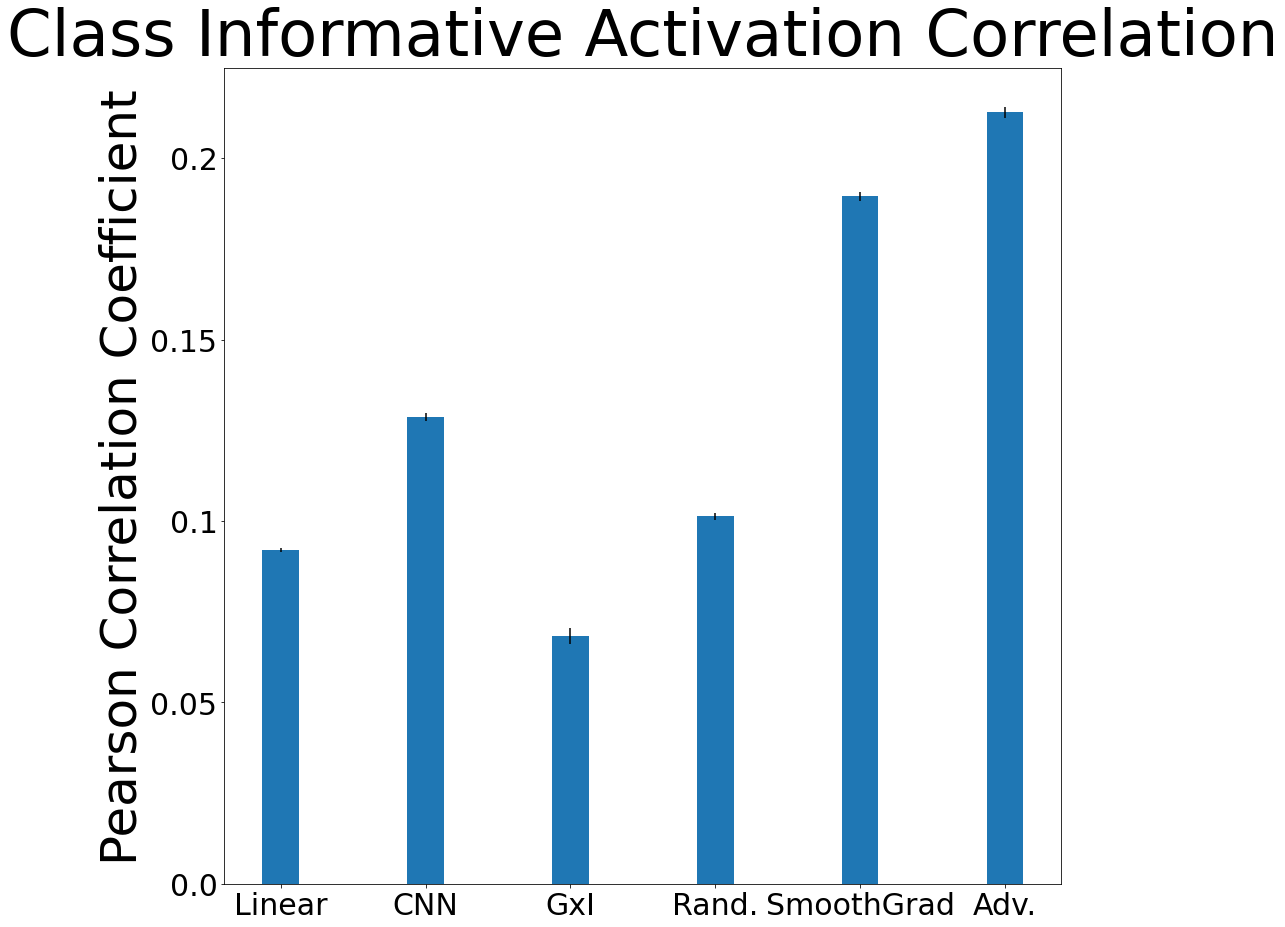}
    \captionsetup{justification=centering}
    \caption{}
        \label{}
  \end{subfigure}
  \begin{subfigure}[b]{0.10\textwidth}
    \centering
    \textnormal{     }
  \end{subfigure}
  \begin{subfigure}[b]{0.4\textwidth}
    \centering
    \includegraphics[width=\textwidth]{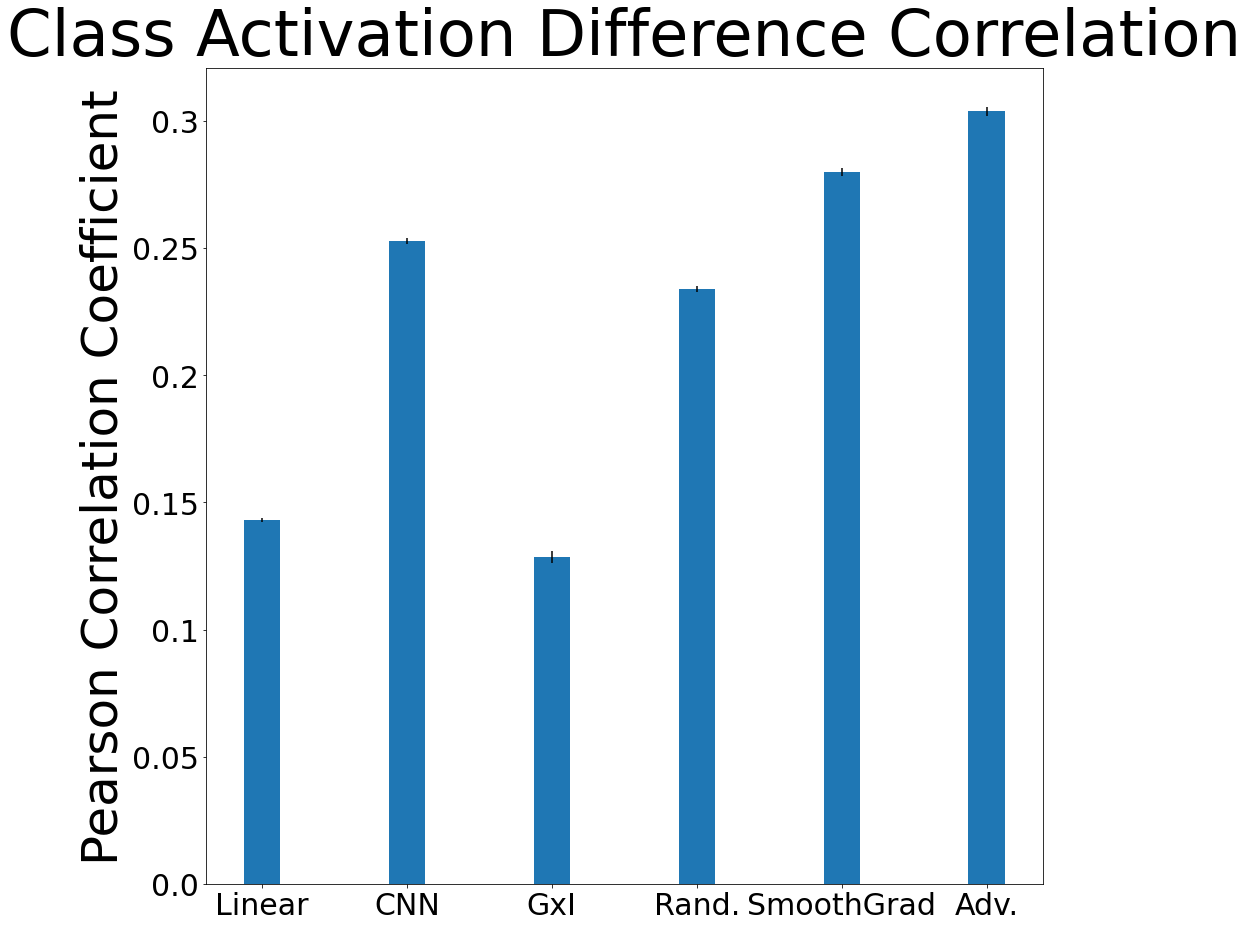}
    \captionsetup{justification=centering}
    \caption{}
  \end{subfigure}
  \caption{{\bf The average interpretability scores (with standard errors) for saliency maps on the HCP test set, according to our two quantitative evaluation criteria.} The scores are calculated for the linear multinomial regression model gradients, convolutional neural network (CNN) gradients, CNN gradients times inputs (GxI), CNN trained with additive random noise (Rand.) gradients, CNN with SmoothGrad gradients, and CNN with adversarial training (Adv.) gradients. {\bf a:} The average correlations versus the task informative activation map, across all test subjects. {\bf b:} The average correlations versus the pairwise class activation difference map, across all tasks and test subjects.} 
  \label{fig:corrs}
\end{figure}

\paragraph{Class Informative Activation}

We computed a ground truth class informative activation map for each decoding class and each subject, by subtracting a subject's average example across classes from the subject's average example of each class.  In order to evaluate each saliency map method, we computed the voxelwise correlation between the map produced for each input example, with the gradient for its correct class, and the class informative activation map. This yields a sample of correlation values for each method; the average correlations for each method are shown in Figure \ref{fig:corrs}a. Per paired t-tests as described above, the CNN with adversarial noise had a statistically higher correlation ($p<0.01$, corrected) than all of the other methods. The linear model had a statistically lower correlation ($p<0.01$, corrected) than all of the other methods, except for the gradient times input method. The gradient times input method had a statistically lower correlation than every other method, including the linear model ($p<0.01$, corrected). Additionally, using SmoothGrad significantly ($p<0.01$, corrected) increased the correlation of the gradients with the task informative activation maps for the CNN trained with and without additive random noise.

\begin{figure}[h]
  \centering
  \begin{subfigure}[b]{\textwidth}
    \centering
    \includegraphics[width=\textwidth]{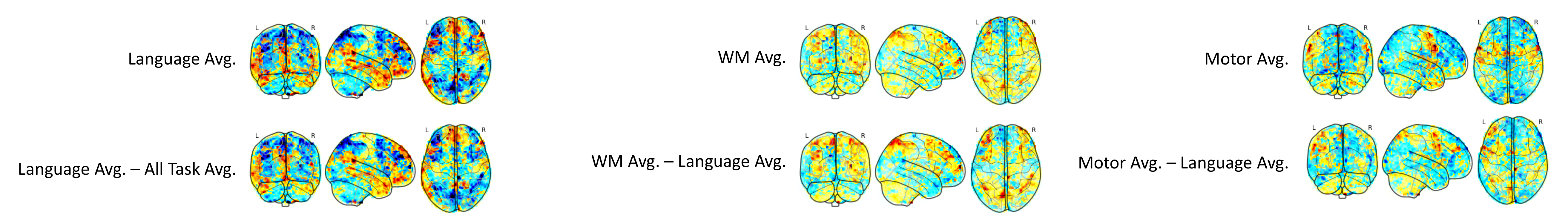}
    \captionsetup{justification=centering}
    \caption{}
  \end{subfigure}
  \begin{subfigure}[b]{\textwidth}
    \centering
    \includegraphics[width=\textwidth]{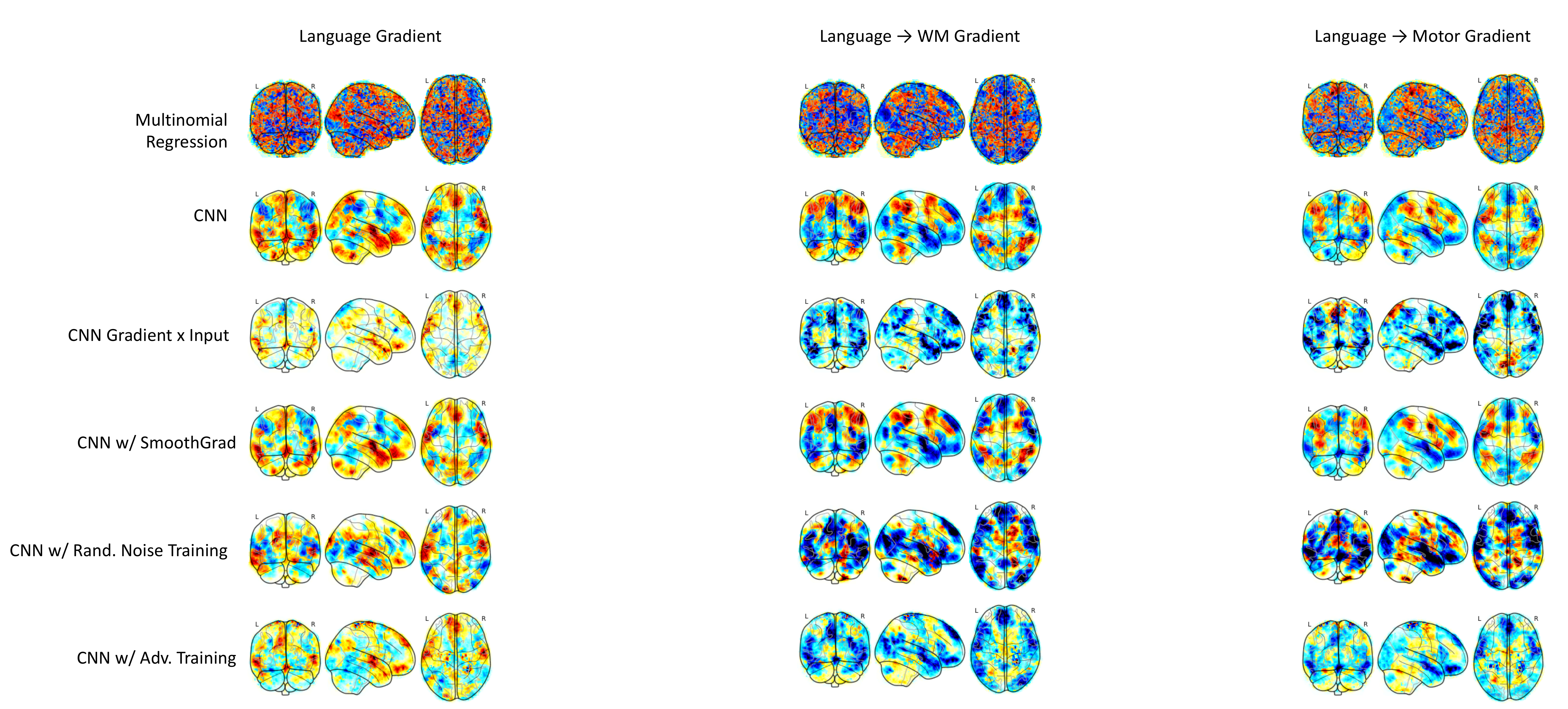}
    \captionsetup{justification=centering}
    \caption{}
  \end{subfigure}
  \caption{{\bf Visualization of gradient-based saliency maps for a test subject from the HCP test data.} {\bf a:} In row 1, the coronal, sagittal, and axial views of "glass brains" of the examples for three classes, Language, Working Memory (WM), and Motor. In row 2, the class informative activation map obtained by subtracting the average example across classes from the Language example (column 1) and the pairwise class activation difference maps for WM (column 2) and Motor (column 3) with respect to Language. {\bf b:} The average gradient-based saliency maps (rows 1-4) for the Language class, applied to the example of the Language (column 1), WM (column 2), and Motor (column 3) classes. (Each "glass brain" was individually scaled, using the 99.9999th percentile of absolute values across the brain to define the extremes.)}
  
  \label{fig:hcp}
\end{figure}

Figure~\ref{fig:hcp}a shows the average example across subjects for three classes (row 1), and the corresponding class informative activation maps (row 2).
The saliency maps produced with the various methods are shown for one class (Language) in column 1 of Figure~\ref{fig:hcp}b.

\paragraph{Pairwise Class Activation Differences}

We computed ground truth class pairwise activation difference maps for each pair of classes and each subject, by subtracting a subject's average activation map, across conditions, for each class from the average maps of all other classes. We applied this approach to the gradient-based saliency maps for each possible class produced for each test example, using each of the evaluated methods.  For each method, and across all test examples, we correlated the pairwise gradient-based saliency maps with the corresponding pairwise class activation difference maps, yielding an average correlation value across incorrect classes; across test examples, this results in a sample of correlation values. For any two methods, we used a two-sided paired t-test to compare whether their average correlation was different. We used Bonferroni correction to account for all of the pairwise comparisons. The average correlations for the different methods are shown in Figure \ref{fig:corrs}b. Per two-sided paired t-tests as described earlier, the CNN with adversarial noise had a higher correlation  ($p <0.01$, corrected)  than  all the  other  method.  The  linear  model had a lower  correlation  ($p <0.01$, corrected) than  all  the  other  methods, except for the gradient times input method. The gradient times input method had a statistically lower correlation than every other method, including the linear model ($p<0.01$, corrected). SmoothGrad was significantly better than the CNN trained with and without additive random noise ($p < 0.01$, corrected).

In Figure~\ref{fig:hcp}b, we show the average gradient-based saliency map produced by each method for the Language class, applied to the average example of the Language (column 1), WM (column 2), and Motor (column 3) classes. Note how these gradient maps remove the most prominent activation in the maps of the correct class (column 1), when transforming to a different class (columns 2 and 3). In particular, note that the gradient times input language to motor gradient map (Figure~\ref{fig:hcp}b, column 3, row 3) removes activation where the input activation is high, but does not indicate that activation should be added around the central sulcus.

\section{Discussion}

DNNs are being increasingly used to make predictions in cognitive neuroscience and neuroinformatics applications. While successful, in terms of prediction performance, they are widely seen as uninterpretable ``black boxes'', as it can be difficult to discover what input information is used to make predictions. A common approach to addressing this concern is to produce gradient-based saliency maps, visualizations of the relative importance of input features in prediction. Many methods for generating saliency maps have been introduced in the machine learning literature. However, they are known to be brittle due to focusing too much on the input, or being extremely sensitive to small input noise. Finally, it is also challenging to evaluate how well saliency maps correspond to the truly relevant input information, given that ground truth is not always available.
There is also no consensus on how to quantitatively evaluate the interpretability of the resulting maps. 

We are seeing increasing usage of saliency map methods in neuroimaging, briefly reviewed earlier in this paper. Hence, we believe it is timely to compare their effectiveness, and do so with a focus on our application domain. To that effect, we introduced two quantitative criteria to evaluate the interpretability of saliency map methods for fMRI task decoding. These criteria are well-defined whenever a model is being trained to decode some information from imaging; they are, therefore, of broad applicability. We first described the rationale for the criteria using a synthetic dataset, where the activation structure is known. There, we also drew the distinction between activation and information (activation present in some tasks but not others), as the latter is what should be captured in a saliency map. In the evaluations of maps in prior neuroimaging work, the ground truth is often a map of known activation for a particular task, so we believe that showing that this distinction matters in practice is a useful contribution.

We provided a brief review of common gradient-based saliency map methods for DNNs: plain gradient, gradient times input, SmoothGrad, gradient trained with random and adversarial noise. We also introduced a new adversarial training method we developed to make DNNs robust to input noise, with the goal of improving interpretability. We then evaluated all of these methods with DNNs trained to perform task decoding, in both synthetic data and the task activation data in the Human Connectome Project (HCP) dataset, using both of our quantitative criteria.

We obtained very consistent results across both synthetic and HCP datasets. First, gradient-based saliency maps produced with different methods vary widely in interpretability, as measured by our evaluation procedures in synthetic and HCP data. Second, even when linear and DNN models decode at comparable levels of performance, saliency maps from the DNN score higher on interpretability than maps derived from the linear model (via weights or gradient). Third, using adversarial training improves the interpretability of CNN gradients relative to {\em all} other methods. Other methods fail by being too sensitive to noise or by introducing information from the input image into the final map (e.g. gradient times input).  Our results suggest that adversarial training is a promising approach when compared with other methods, and can be carried out in a computationally efficient manner with the procedure we introduce. 

With regards to future work, the most obvious direction is to expand this comparison to other large datasets with many more tasks, as the distinction between activation and information will become even more critical. In particular, we could apply adversarial training procedure to regression problems such as trait and measure prediction problems, which are likely to be more clinically relevant. Finally, we would like to to address the limitations of using gradients, which are first-order (i.e. linear), local approximations of the optimization landscape, by using input Hessians (i.e. second-order approximations of the optimization landscape). This should increase the interpretability of complex machine learning models and their widespread adoption in the scientific community.

\section*{Acknowledgements}
This work utilized the computational resources of the NIH HPC Biowulf cluster (http://hpc.nih.gov). Data were provided in part by the Human Connectome Project, WU-Minn Consortium (Principal Investigators: David Van Essen and Kamil Ugurbil; 1U54MH091657) funded by the 16 NIH Institutes and Centers that support the NIH Blueprint for Neuroscience Research; and by the McDonnell Center for Systems Neuroscience at Washington University.


\bibliographystyle{vancouver}
\bibliography{bib}

\end{document}